# End-to-end Training for Whole Image Breast Cancer Diagnosis using An All Convolutional Design

Li Shen
Icahn School of Medicine at Mount Sinai
1425 Madison Ave, New York, New York, USA
li.shen@mssm.edu

## Abstract

We develop an end-to-end training algorithm for whole-image breast cancer diagnosis based on mammograms. It requires lesion annotations only at the first stage of training. After that, a whole image classifier can be trained using only image level labels. This greatly reduced the reliance on lesion annotations. Our approach is implemented using an all convolutional design that is simple yet provides superior performance in comparison with the previous methods. On DDSM, our best single-model achieves a per-image AUC score of 0.88 and three-model averaging increases the score to 0.91. On INbreast, our best single-model achieves a per-image AUC score of 0.96. Using DDSM as benchmark, our models compare favorably with the current state-of-the-art. We also demonstrate that a whole image model trained on DDSM can be easily transferred to INbreast without using its lesion annotations and using only a small amount of training data. Code availability: https://github.com/lishen/end2end-all-conv

## 1    Introduction

With the rapid advancement in machine learning and especially, deep learning in recent years, there is a keen interest in the medical imaging community to apply these techniques to cancer screening. Mammography based breast cancer diagnosis is a challenging problem in which whole-image level labels are determined by small, localized regions that have huge variations among patients. Down-sampling a large mammogram to the typical size of 224x224 for image classification will likely make the regions of interest (ROIs) hard to detect and/or classify. If a mammographic database comes with the ROI annotations for all images, then the diagnostic problem can be conveniently solved as an object detection and classification problem that has already been well studied in the computer vision field. For example, the region-based convolutional neural network (R-CNN) approach [1] and its variants [2]–[4] can be applied here. Many of the published works [5]–[14] assume the databases under study are fully annotated with ROIs, thus the developed models cannot be transferred to another database that lacks ROI annotations. It is unrealistic to require all databases to be fully annotated due to the time and monetary costs to obtain ROI annotations, which require expertise from radiologists. For a survey of the public databases, see [15]. If we want to build a breast cancer diagnostic system for production, we will have to consider the situation where a few datasets are annotated with ROIs while most datasets are only annotated at whole image level.

In this study, we propose an approach that utilizes a fully annotated database to first train a model to recognize localized patches. After that, the model can be converted into a whole image classifier, which can be trained end-to-end without ROI annotations. The approach is based on an idea that we independently formed (not implemented due to time limit) during the digital mammography (DM) challenge (see discussion in [16]), which coincides with the approach [17] that won the challenge. Here, we propose an all convolutional design that improves upon the winning strategy. It is simple and elegant and yet provides very competitive results. We also present a pipeline to build a whole image classifier from scratch and discuss the pros and cons of some important choices.

## 2    Methods

### 2.1    Patch to image conversion

A convolutional neural network (CNN) is recursively defined so that when a new convolutional layer is added on the top, it effectively uses all previous convolutional layers to construct a set of new filters that perform more complex transformations than its precursors. By this reasoning, we can add on top of a pretrained patch classification network a new convolutional layer and turns the patch network itself into part of the new filters of the top layer. In order to make this new network detect and classify abnormalities on the whole image, we just need to modify the input of the new network from patches to whole images. This way, the patch classifier can be used to efficiently "scan" the whole image in one forward propagation; instead of generating one prediction for a single patch, it will generate predictions for all overlapping patches on the image, creating a so called "heatmap" that represents the likelihood of each patch being background, benign or malignant. The top layers can then be trained to learn from the outputs of the patch classifier and connect with the image-level labels. See Appendix 1 for an illustration of the whole process. Notice that variable input size can be easily implemented in most deep learning frameworks [18]–[21].

## 2.2 Network structures for top layers

For top layers, we propose an all convolutional design that makes no use of the fully connected (FC) layers. Briefly, we remove the last few layers of a patch network until the last convolutional layer and then add more convolutional layers on top to reduce feature map sizes. This continues until the feature map reaches a small size and then global average pooling is applied to the final convolutional layer's output to connect with the image-level output. Notice that we deliberately discard the "heatmap" layer because it has a very small dimension, such as 3, which may create an "information barrier" that prevents the top layers from fully utilizing the information provided by the patch network.

In contrast to our approach, the top performing team of the DM challenge utilized the heatmap [17]. They also made use of FC layers to connect the heatmap with the image-level output. We argue that FC layers are not very good for image recognition because they eliminate all the spatial information. Our design is fully convolutional and preserves spatial information at every layer (except the last) of the network. We will compare the two different strategies in the following sections.

## 2.3 Network training

There are two parts in training a whole image classifier from scratch. The first part is to train a patch classification network. We use the networks that have their weights pretrained on the ImageNet [22] database and expect the bottom layers to be more preserved than the top layers. Therefore, we develop a 3-stage training strategy as follows:

1. Set learning rate to 1e-3 and train only the last layer for 3 epochs.
2. Set learning rate to 1e-4, unfreeze the top layers and train for 10 epochs.
3. Set learning rate to 1e-5, unfreeze all layers and train for 37 epochs.

The top layer number can be arbitrarily chosen to represent the high-level features and shall vary according to different network structures. The number of epochs for each stage is arbitrarily chosen so that the total number of epochs is 50, which is mainly decided by the available computational resource. Adam [23] is used as the optimizer and the batch size is set to 32. We also adjust the sample weights within a batch to keep the classes balanced.

The second part is to convert the patch classifier into a whole image classifier. By altering the input size from patch to whole image, we proportionally increase the feature map size for every convolutional layer. We then add top layers to connect with the output as discussed in Section 2.2. Similar to the patch network training, we develop a 2-stage training strategy to learn the top layers first to avoid unlearning the important features in the patch network. Due to memory constraint, we use a small batch size of 2 for whole image training.

We calculate the pixel-wise mean for the images on the train set and use this value for pixel-wise mean centering for both patch and whole image training. No other preprocessing is applied. To compensate for the lack of sample size, we use the same data augmentation on-the-fly for both patch and whole image training.

## 3 Results

### 3.1 Classifier training on DDSM

#### 3.1.1 Setup and processing of the dataset

We use a modernized version of the Digital Database for Screening Mammography (DDSM) called CBIS-DDSM [24] which contains images in DICOM format. The downloaded (Date: 3/27/2017) dataset contains 1249 patients or 2584 mammograms, which represents a curated subset of the DDSM. It includes both cranial cardo (CC) and media lateral oblique (MLO) views but we will treat each view as a standalone image in this study. We perform an 85-15 stratified split on the dataset into train and test sets based on the patients and their cancer status. For the train set, we further set aside 10% of the patients as the validation set. The total numbers of images in the train, validation and test sets are: 1903, 199 and 376, respectively.

The DDSM database contains the pixel-level annotation for the ROIs and their pathology confirmed labeling: benign or malignant and also their type: calcification or mass. The numbers of images of the four combinations of label and type are about the same. We create a patch image set by sampling 10 patches from each ROI with a minimum overlapping ratio of 0.9 and the same number of background patches from the same mammogram. According to the annotations of the ROIs, a patch is classified into five categories: background, calcification-benign, calcification-malignant, mass-benign and mass-malignant.

#### 3.1.2 Development of image classifiers

Table 1: Test accuracy of the patch classifiers using the Resnet50 and VGG16. #Epochs indicates the epoch when the best validation accuracy has been reached.

| Model | Accuracy | #Epochs |
|---|---|---|
| **Resnet50** | 0.89 | 39 |
| **VGG16** | 0.84 | 25 |

To train a network to classify a patch into the five categories, we use two popular convolutional network structures: the Resnet50 [25] and the VGG16 [26]. Because the number of background patches is matched with that of ROI patches on the same mammogram, the five categories are roughly balanced. Therefore, we evaluate the models using test accuracy and find that both networks achieve high accuracy while Resnet50 outperforms VGG16 by 0.05 (Table 1). We now convert the patch classifiers into whole image classifiers by testing many different configurations for the top layers. We evaluate different models using the per-image receiver operating characteristic curve (AUC) scores on the test set. We first test the conversion based on the Resnet50 patch classifiers (Table 2). In our first test, we use two residual blocks with the same dimension and a bottleneck design (see [25]) of [512-512-2048] without repeating the residual units. This gives an AUC score of 0.85. We further vary the design of the residual blocks by reducing the dimension of the last layer in each residual unit to 1024, which allows us to repeat each residual unit twice without exceeding memory constraint. This design slightly increases the score to 0.86. We also reduce the dimensions of the first and second blocks and find the scores to drop by only 0.02. Therefore, we conclude that the dimensions for the newly added residual blocks are not critical to the performance of the whole image classifiers.

Table 2: Per-image test AUC scores of the whole image classifiers using Resnet50 and VGG16 as patch classifiers. #Epochs indicates the epoch when the best validation score has been reached. Each residual block is represented by [M-N-O]xR, where M=dimension of the 1st 1x1 conv layer; N=dimension of the 2nd 3x3 conv layer; O=dimension of the 3rd 1x1 conv layer; R=number of repetitions of the unit. Each VGG block is represented by MxR, where M=dimension of the 3x3 conv layer; R=number of repetitions of the unit.

| Patch net | Block1 | Block2 | Single-model AUC | Augmented AUC | #Epochs |
|---|---|---|---|---|---|
| **All convolutional design** | | | | | |
| Resnet50 | [512-512-2048]x1 | [512-512-2048]x1 | 0.85 | NA | 20 |
| Resnet50 | [512-512-1024]x2 | [512-512-1024]x2 | 0.86 | 0.88 | 25 |
| Resnet50 | [256-256-512]x3 | [128-128-256]x3 | 0.84 | NA | 48 |
| VGG16 | 512x3 | 512x3 | 0.71 | NA | 47 |
| VGG16 | 512x1 | 512x1 | 0.83 | 0.86 | 44 |
| VGG16 | 256x1 | 128x1 | 0.80 | NA | 35 |
| VGG16 | [512-512-1024]x2 | [512-512-1024]x2 | 0.81, 0.85[1] | 0.88[1] | 46 |
| **Add heatmap and residual blocks on top** | | | | | |
| Resnet50 | [512-512-1024]x2 | [512-512-1024]x2 | 0.80 | NA | 47 |
| **Add heatmap, max pooling and FC layers on top** | | | | | |
| | Pool size | FC1　FC2 | | | |
| Resnet50 | 5x5 | 64　32 | 0.73 | NA | 28 |
| VGG16 | 5x5 | 64　32 | 0.71 | NA | 26 |

[1]Result obtained from extended model training (See text for more details).

We then test the conversion based on the VGG16 patch classifier (Table 2). The newly added VGG blocks all use 3x3 convolutions with batch normalization (BN). We find that the VGG structure is more likely to suffer from overfitting than the residual structure (see Appendix 2): add two 512x3 blocks achieves a score of only 0.71. However, this can be alleviated by reducing the model complexity of the newly added VGG blocks: add two 512x1 blocks increases the score to 0.83. We further reduce the dimensions of the VGG blocks and find the scores to decrease by only 0.03. This is in line with the results on the Resnet50 based models. Overall, the Resnet50 based whole image classifiers perform better than the VGG16 based ones (Table 2). In addition, the Resnet50 based models seem to achieve the best validation score earlier than the VGG16 based models. To understand whether this performance difference is caused by the patch network on the bottom or the newly added top layers, we add two residual blocks on top of the VGG16 patch network to create a "hybrid" model. It gives a score of 0.81, which is in line with the VGG16 based networks but a few points below the best Resnet50 based models. This suggests the patch network part is more important than the top layers to whole image classification performance. However, after extended training, this hybrid model achieves a score of 0.85 (also see Section 3.2.2).

Direct comparison of our method with that of the top performing team [17] is not allowed at the moment[1]. Therefore, we use a design that is inspired by the winning method: adding a heatmap followed by max pooling and two FC layers and a shortcut connection. Obviously, image classifiers from this design underperform the ones from our all convolutional design by as much as 15 points (Table 2). Notice that in [17], a modified VGG network of 6 instead of 5 VGG blocks was used. Their design is more like a hybrid of the two designs compared here by using an additional convolutional block.

---

[1] When this manuscript was written, the DREAM challenge team had not published their results yet. According to challenge policy, no participant is allowed to publicly discuss the result of a method from the challenge.

We also want to find out whether the heatmaps are of any use. We add a heatmap on top of the Resnet50 patch classifier and then add two residual blocks with [512-512-1024]x2 design. This network gives a score of 0.80 (Table 2), which is lower than the same design without the heatmap. We conclude that the heatmap shall be removed to facilitate information flow in training whole image networks.

### 3.1.3 Model averaging

We pick a few models and use inference-level augmentation by doing horizontal and vertical flips to create four predictions per model and take an average. Three best performing models: Resnet50 with two [512-512-1024]x2 residual blocks, VGG16 with two 512x1 VGG blocks and VGG16 with two [512-512-1024]x2 residual blocks on top are selected. The augmented predictions for the three models improve the AUC scores from 0.86→**0.88**, 0.83→**0.86** and 0.85→**0.88**, respectively (Table 2). The average of the three augmented predictions gives an AUC score of **0.91**. Notice that we deliberately avoid doing too much model averaging so that we can better understand the pros and cons of different network structures.

## 3.2 Transfer learning on INbreast

### 3.2.1 Setup and processing of the dataset

The INbreast [15] dataset is another public database for mammograms, which contains full-field digital images. These images have different color profiles from the images of DDSM (see Appendix 3). Therefore, this is an excellent source to test the transferability of a whole image classifier from one database to another. The INbreast database contains 115 patients and 410 mammograms and includes the BI-RADS readings for the images. We manually assign all images with BI-RADS readings of 1 and 2 as negative samples; 4, 5 and 6 as positive samples; and ignore BI-RADS readings of 3. This excludes 12 patients or 23 mammograms from further analysis. We perform a 70-30 split on the dataset into train and validation sets based on the patients in a stratified fashion. The total numbers of images in the train and validation sets are 280 and 107, respectively. We use the same processing steps on INbreast images as the DDSM.

### 3.2.2 Efficiency of transfer learning

We directly finetune the whole image networks on the train set without using ROI annotations and evaluate the model performance using per-image validation AUC scores. The three best models from Section 3.1.3 are used for transfer learning. The finetuned Resnet50 based model achieves a score of 0.84. Surprisingly, the finetuned VGG16 based model achieves a score of 0.92, better than the Resnet50 based models. It is argued in [17] that the residual networks reduce the feature map sizes too aggressively to damage the ROI features at the first few layers. Based on this argument, the underperformance of the Resnet50 based models is likely due to the bottom most layers. This is validated by the high performance of the hybrid model, which achieves a score of **0.95**. Therefore, we hypothesize that the VGG networks need to be trained longer than the residual networks to reach their full potentials on the DDSM. To prove that, we perform another run of model training for the hybrid model with 200 additional epochs. The model improves the test score from 0.81 to 0.85 (Table 2), which is as good as the best Resnet50 based models.

We also want to find out how much data is required to finetune a whole image classifier to reach satisfactory performance. This has important implications in practice since obtaining labels, even at the whole image level, can be expensive. We sample a subset with 20, …, 60 patients from the train set for finetuning and evaluate the model performance on the same validation set (Table 3). With as little as 20 patients or 79 images, the VGG16 based model and the hybrid model can achieve scores of 0.87 and 0.89, respectively. The scores seem to quickly saturate as we increase the train set size. This quick adjustment can be a huge advantage for the end-to-end trained whole image networks, which greatly reduce the burden of train set construction.

Table 3: Transfer learning efficiency with different train set sizes. Shown are per-image validation AUC scores.

| #Pat | #Img | Resnet50 | VGG16 | Hybrid |
|------|------|----------|-------|--------|
| 20   | 79   | 0.78     | 0.87  | 0.89   |
| 30   | 117  | 0.78     | 0.90  | 0.90   |
| 40   | 159  | 0.82     | 0.90  | 0.93   |
| 50   | 199  | 0.80     | 0.93  | 0.93   |
| 60   | 239  | 0.84     | 0.95  | 0.91   |

Finally, with augmented prediction, the VGG16 based model improves the score from 0.92→0.94 and the hybrid model improves the score from 0.95→**0.96**. The average of the two augmented models gives a score of **0.96**. Including the Resnet50 based model in model averaging does not improve the score.

**Appendix 1: Patch to whole image conversion**

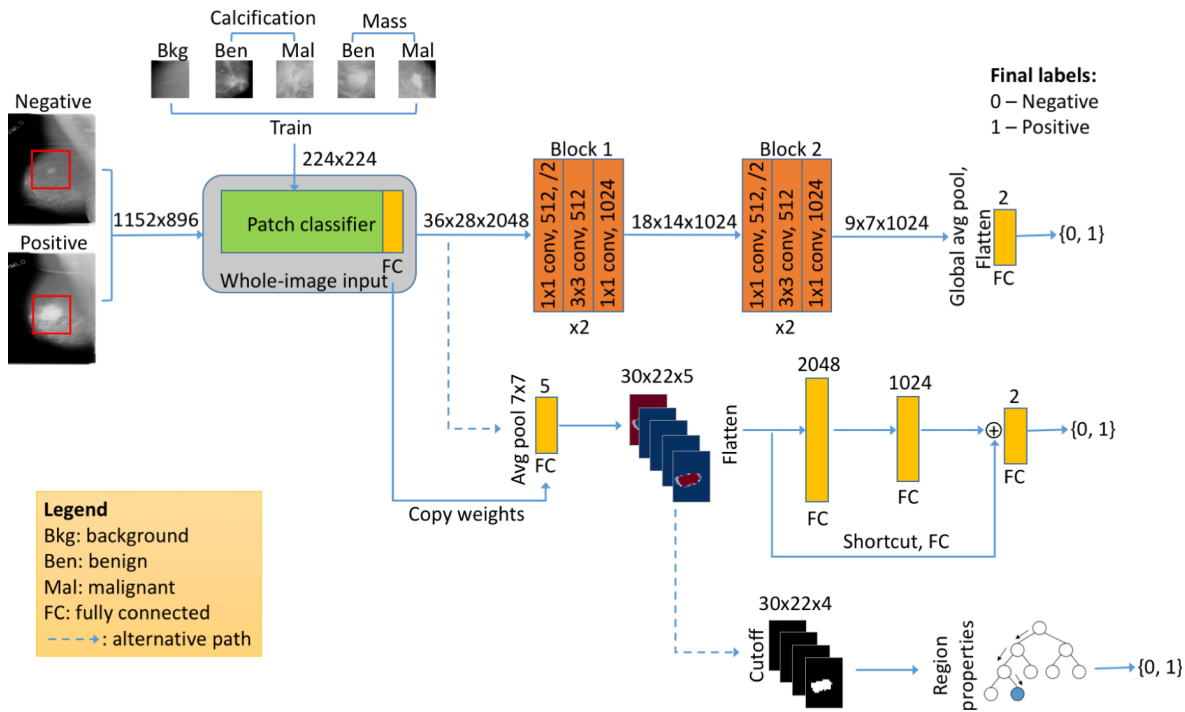

The deep learning structure for converting a patch classifier into a whole image classifier by adding convolutional layers on top. Shown is an example using two residual blocks with the same structure of [512-512-1024] x 2. Two alternative strategies are also presented: one is to add heatmaps and FC layers on top and the other is to add a random forest classifier on top of the heatmap.

**Appendix 2: Overfitting of VGG structures**

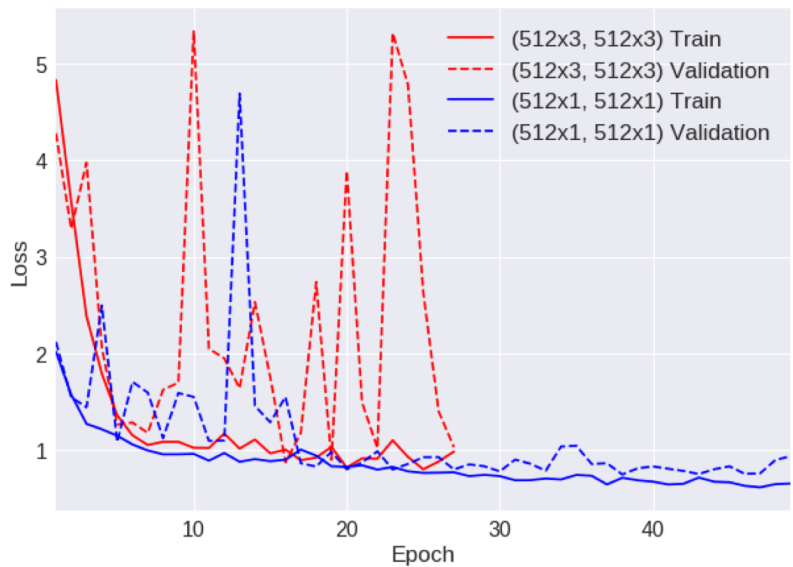

Train and validation loss curves of two VGG structures: one is more complex than the other and suffers from overfitting.

**Appendix 3: the DDSM and INbreast data have different color profiles as illustrated by two example mammograms**

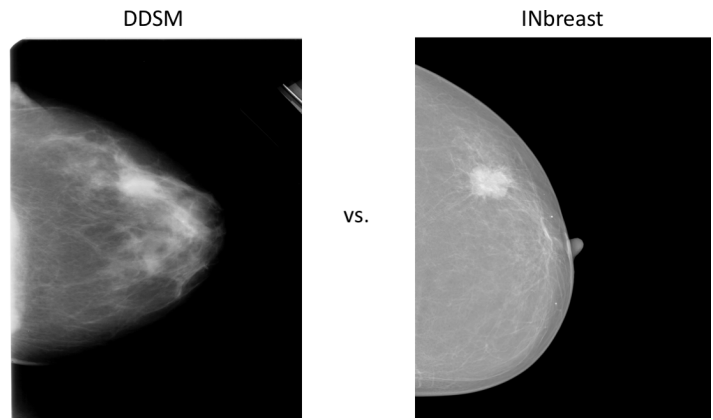

Comparison of two example mammograms from DDSM and INbreast.